\documentclass{article}

   \usepackage[preprint]{neurips_2021}

\usepackage[utf8]{inputenc} 
\usepackage[T1]{fontenc}    
\usepackage{hyperref}       
\usepackage{url}            
\usepackage{booktabs}       
\usepackage{amsfonts} 
\usepackage{nicefrac}       
\usepackage{microtype}      
\usepackage{xcolor}         
\usepackage{graphicx}
\usepackage{fix-cm}
\usepackage{amssymb}
\usepackage{multirow}
\usepackage{multicol}
\usepackage{ccaption}
\usepackage{subcaption}
\usepackage{rotating}
\usepackage{stackrel}
\title{ Task Oriented Conversational Modelling With Subjective Knowledge}

\author{Raja Kumar\\ \\
University of California Santa Cruz\\
\texttt{rkumar38@ucsc.edu}}

\begin{document}

\maketitle

\begin{abstract}
Existing conversational models are handled by a database(DB) and API based systems. However, very often users' questions require information that cannot be handled by such systems. Nonetheless, answers to these questions are available in the form of customer reviews and FAQs. DSTC-11 proposes a three stage pipeline consisting of knowledge seeking turn detection, knowledge selection and response generation to create a conversational model grounded on this subjective knowledge. In this paper, we focus on improving the knowledge selection module to enhance the overall system performance. In particular, we propose entity retrieval methods which result in an accurate and faster knowledge search. Our proposed Named Entity Recognition (NER) based entity retrieval method results in 7X faster search compared to the baseline model. Additionally, we also explore a potential keyword extraction method which can improve the accuracy of knowledge selection. Preliminary results show a 4 \% improvement in exact match score on knowledge selection task. The code is available \href{https://github.com/raja-kumar/knowledge-grounded-TODS}{here}
\end{abstract}

\section{Introduction}
\begin{figure}
\centering
\includegraphics[scale=0.6]{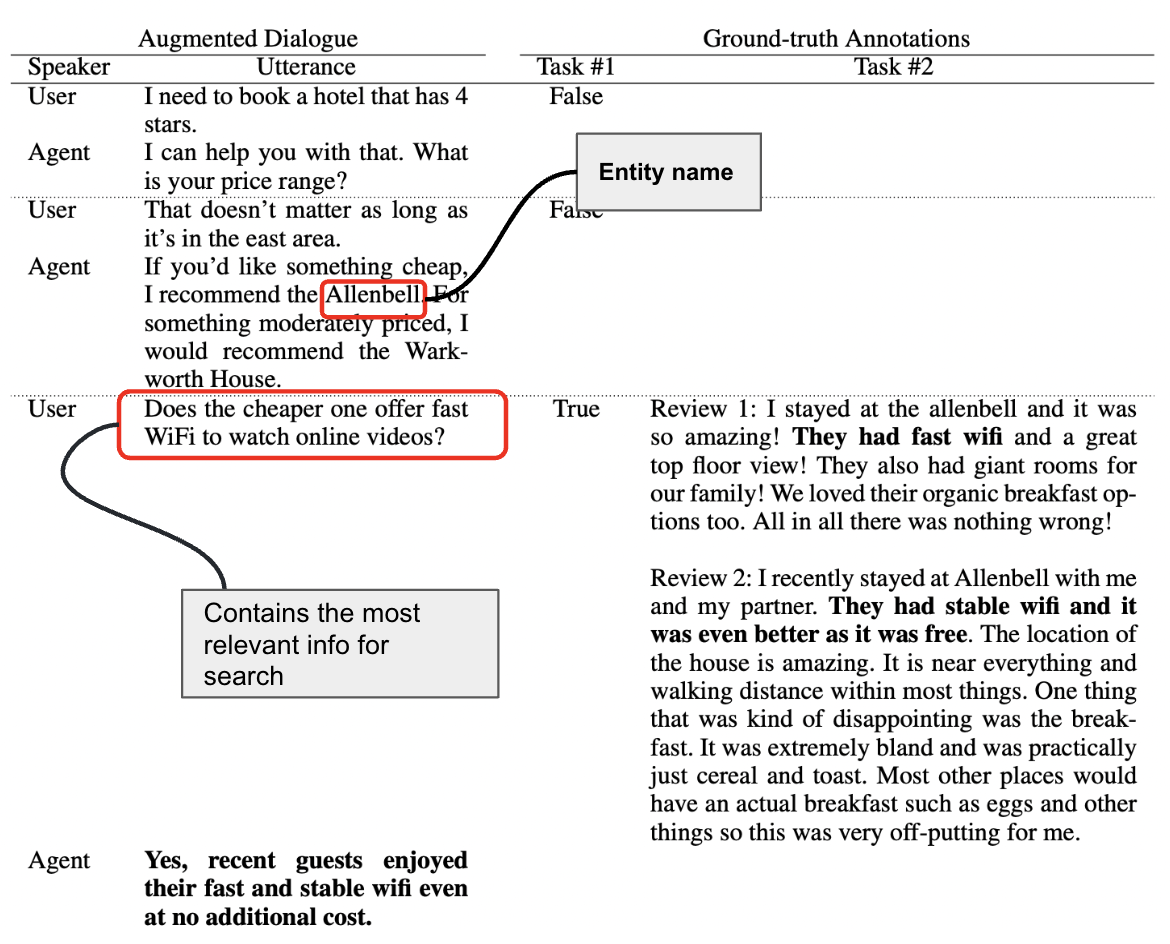}
\caption{An Example Conversation with Subjective Knowledge}
\label{fig:figure1}
\end{figure}

With the advent of Artificial Intelligence (AI), conversational models, also referred to as dialogue systems interchangeably, have become the need of many industries ranging from banking, medical to customer services. Most of these dialogue systems are built on domain specific databases (DB) and APIs \citep{kim-etal-2020-beyond}. It converts user input to a semantic representation such as domain, intent, and slot, which is then used by the back-end to create a response. However, such DB based models either give unnatural responses or fail to answer specific user queries for which data is unavailable \citep{kim-etal-2020-beyond}. Nonetheless, this information is available in the form of customer reviews or FAQs. An example conversation as shown in Fig. \ref{fig:figure1} illustrates such an instance. It can be seen in the example conversation that the last question about whether the hotel offers WiFi to watch online videos cannot be handled by the existing DB. However, there are customer reviews which provide this information. Therefore, a need arises to build a dialogue system which can be grounded on the available knowledge so that the response generated based on this knowledge could be much more engaging and accurate. In addition to building an engaging and cohesive conversation, the system should also be fast and efficient as these systems are used in real-time by users.

\begin{figure}
\centering
\includegraphics[scale=0.4]{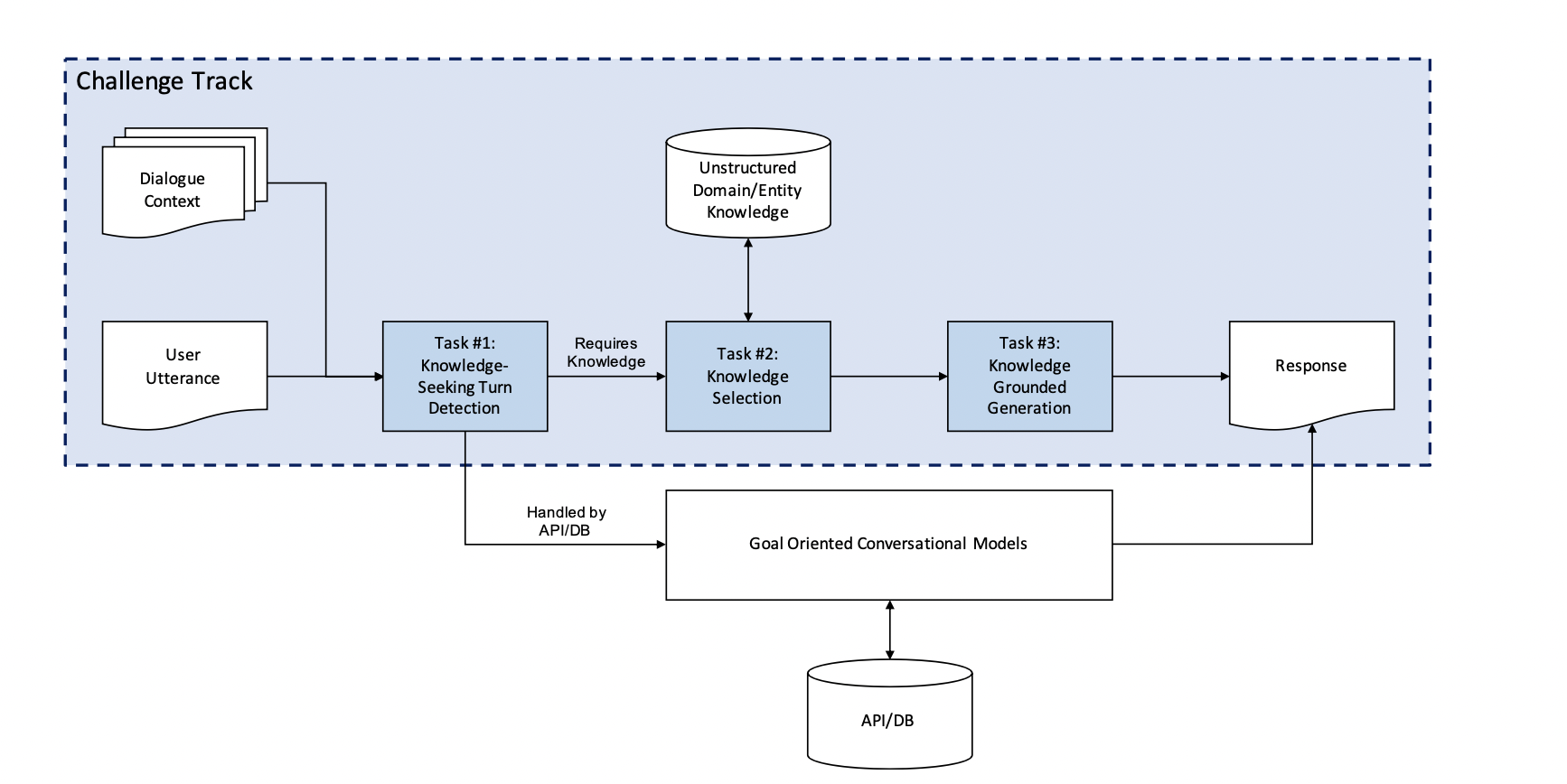}
\caption{An overview of the three stage pipeline proposed by dstc-11 \citep{dstc11}}
\label{fig:figure2}
\end{figure}

DSTC-11 \citep{dstc11} proposes a three stage pipeline consisting of knowledge seeking turn detection, knowledge selection and response generation (see fig \ref{fig:figure2}). Given the conversation history and user utterance, knowledge seeking turn detection (KSTD) step detects whether there is a need for knowledge from the knowledge source or it can be handled by the DB? If  KSTD detects a need for knowledge, then Knowledge selection is performed. In Knowledge selection (KS), the model finds the knowledge which are relevant to the current query and conversation history. Finally the response generation (RG) model generates the response based on the selected knowledge and the conversation history.

\begin{figure}[h]
\centering
\includegraphics[trim = 1.2cm 15cm 1.6cm 2cm, clip=true, width=\linewidth]{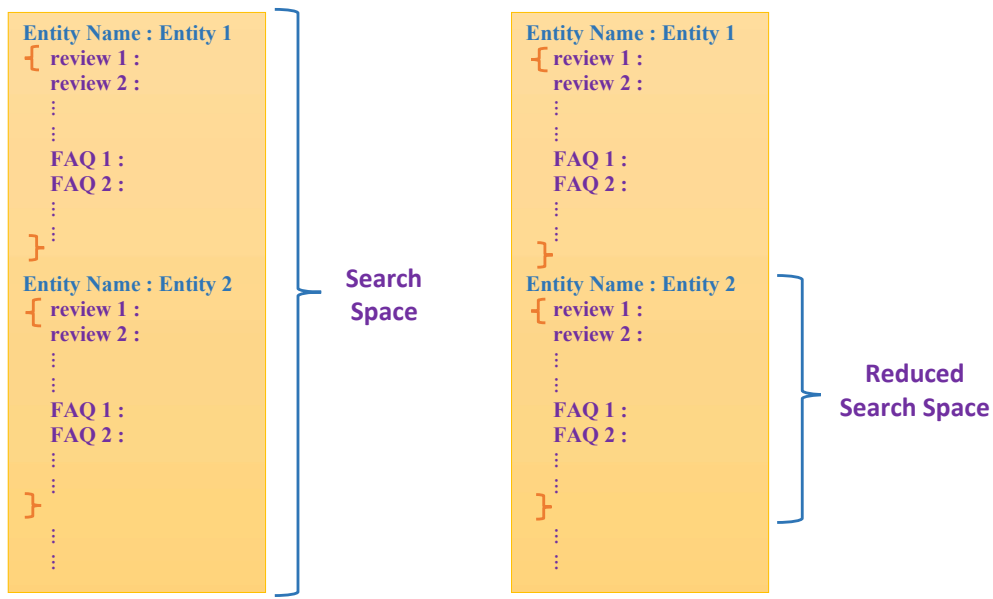}
\caption{An overview of how entity retrieval can help reduce the search space - (a) search space when we don't know the entity name (left) (b) reduced search space once we find the entity name as "Entity 2" (right).}
\label{fig:figure3}
\end{figure}

\begin{figure}
\centering
\includegraphics[trim = 1.2cm 16.5cm 5.5cm 2cm, clip=true, width=\linewidth]{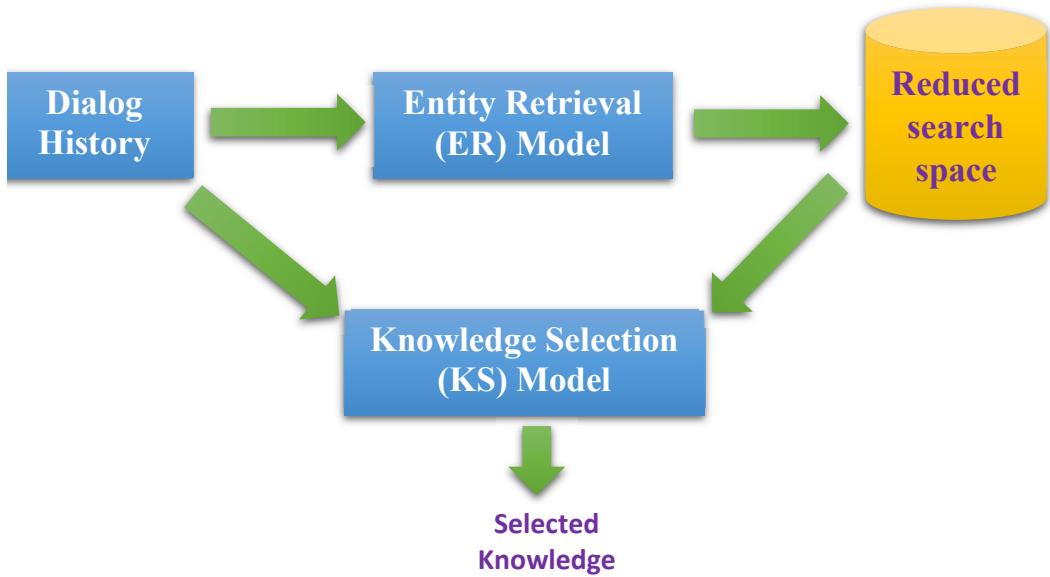}
\caption{The proposed two stage knowledge selection pipeline}
\label{fig:figure4}
\end{figure}

In this paper, We further break down the knowledge selection into a two steps consisting of entity retrieval and knowledge matching (see \ref{fig:figure4}). During the entity retrieval step, the entity name is extracted from the dialogue history which results in a much smaller search space compared to the baseline model (see figure \ref{fig:figure3}). Once we find the entity, we perform knowledge matching on the reduced knowledge base to find the top knowledge candidates. For entity retrieval, we propose two methods. First, we propose entity retrieval using fuzzy logic which retrieves the correct entity with 100\% accuracy on validation data. Second we propose entity retrieval using Named Entity Recognition (NER) which is 7x faster than the proposed fuzzy logic based entity retrieval method. For the knowledge matching, we first perform keyword extraction and then perform knowledge similarity to find the top candidates.

The rest of the paper is organized as follows. In Section \ref{prior work}, a brief review on the related work in dialogue systems is presented and the knowledge grounded conversational model is discussed. The proposed method is described in Section \ref{proposed method} followed by the implementation and training details in Section \ref{implementation details}. Finally, the results obtained using the proposed two-stage knowledge selection method are presented in Section \ref{results} followed by a brief discussion on the future scope of this work and conclusions in Sections \ref{future work} and \ref{conclusion} respectively.

\section{Prior Work}\label{prior work}

Several studies have explored ways to effectively utilize available knowledge base in a task-oriented dialogue(TOD) system. \citep{hosseini2020simple} uses a single language model trained on all sub-tasks through a simple TOD system to improve the performance of dialogue state-tracking through GPT2. \citep{wu2020tod} proposes TOD-BERT, which integrates user and system tokens into language modeling, and outperforms BERT in intent recognition, dialogue state tracking, dialogue act prediction, and response selection by training with ToD datasets. While some models generate conversational interactions based on a general scenario, \citep{ghazvininejad2018knowledge} utilizes dialogue context and knowledge documents as an encoder-decoder model. Additionally, \citep{zhao2020knowledge} efficiently generates responses based on unlabeled dialogues by optimizing knowledge selection through unsupervised learning in relevant knowledge documents.

Various Information Retrieval (IR) techniques have also been used in literature to incorporate the desired knowledge, and successfully implement a knowledge-grounded system. \citep{song2018ensemble} extracts keywords from a query dataset using Term Frequency–Inverse Document Frequency (TF-IDF) to search for the most relevant reply, while \citep{yan2018coupled} generates relevant information for user responses. A document-grounded matching network is presented in \citep{zhao2019document} to use external knowledge in selecting responses for a knowledge-aware retrieval-based chatbot system, that achieves state-of-the-art performance. In addition, an end-to-end process is proposed in \citep{gu2020filtering} that directly learns ranking scores using neural networks. Further, knowledge generation and generation based on a pre-trained language model have been successfully implemented in \citep{zhao2020knowledge}.

\section{Proposed Method}\label{proposed method}

This section discusses the proposed method in detail. Figure \ref{fig:figure3} shows an instance where if we can find the right entity and keywords, we can make our selection model faster and more accurate. Based on this intuition, Figure \ref{fig:figure4} shows the proposed two step knowledge selection model consisting of entity retrieval and knowledge matching. This report proposes two variants of entity retrieval methods - (a) entity retrieval using fuzzy logic and, (b) entity retrieval using NER.

\subsection {Entity retrieval using fuzzy logic}

\begin{figure}[h]
\centering
\includegraphics[scale=0.5]{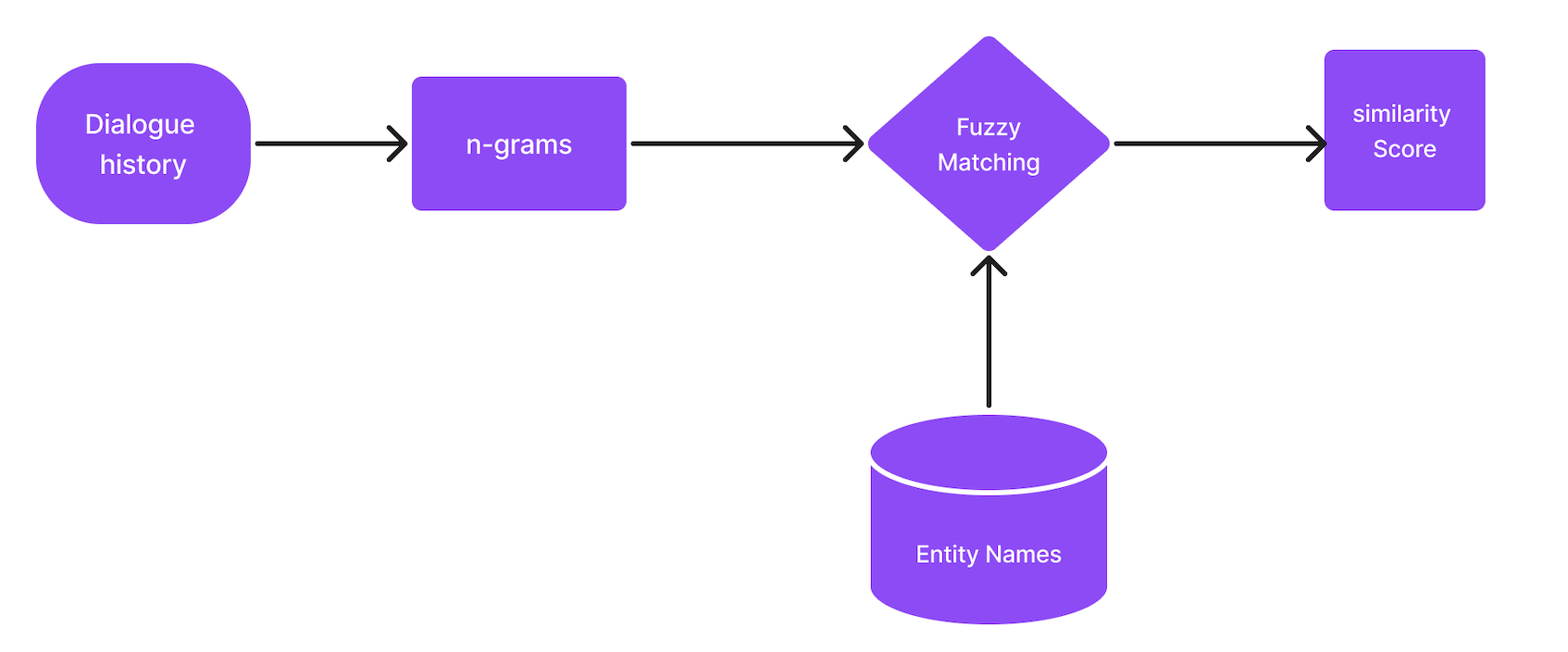}
\caption{The proposed Entity retrieval using fuzzy logic}
\label{fig:figure5}
\end{figure}

In the proposed entity retrieval method using fuzzy logic, we find the entity by comparing all the entity names with the n-grams created using the conversation history. Figure \ref{fig:figure5} shows the proposed methodology. We first convert the conversation history to $n$-grams, where $n$ is the maximum length of the entity. Then, we perform the fuzzy matching between these $n$-grams with all the entities present in the knowledge database and select the one with the highest similarity score as the final entity name. We put a threshold score of $0.95$ to find only the relevant entities.

Using the proposed method, we are able to find the correct entity $100\%$ of the time, on validation data. However, this method is still slow as it requires comparing the similarity between all the $n$-grams and entity names. Therefore, we propose the NER based entity recognition which extracts the entity name directly from the conversation history without the need to compare with all the entity names.

\subsection {Entity retrieval using NER}

\begin{figure}
\centering
\includegraphics[scale=0.4]{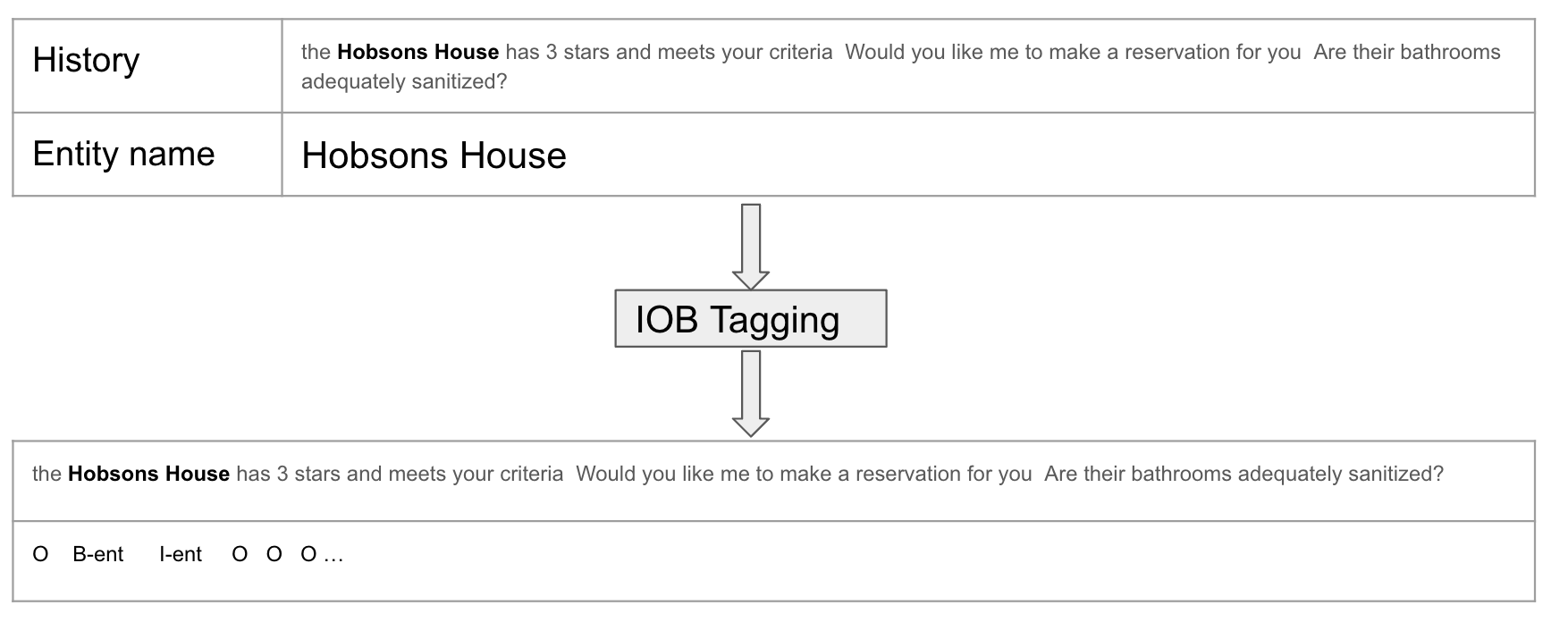}
\caption{A sample example for IOB tagging}
\label{fig:figure6}
\end{figure}

\begin{figure}
\centering
\begin{subfigure}[b]{0.5\textwidth}
\includegraphics[trim = 0.1cm 0.1cm 0.1cm 0.5cm, clip=true,width=7cm,height=4cm]{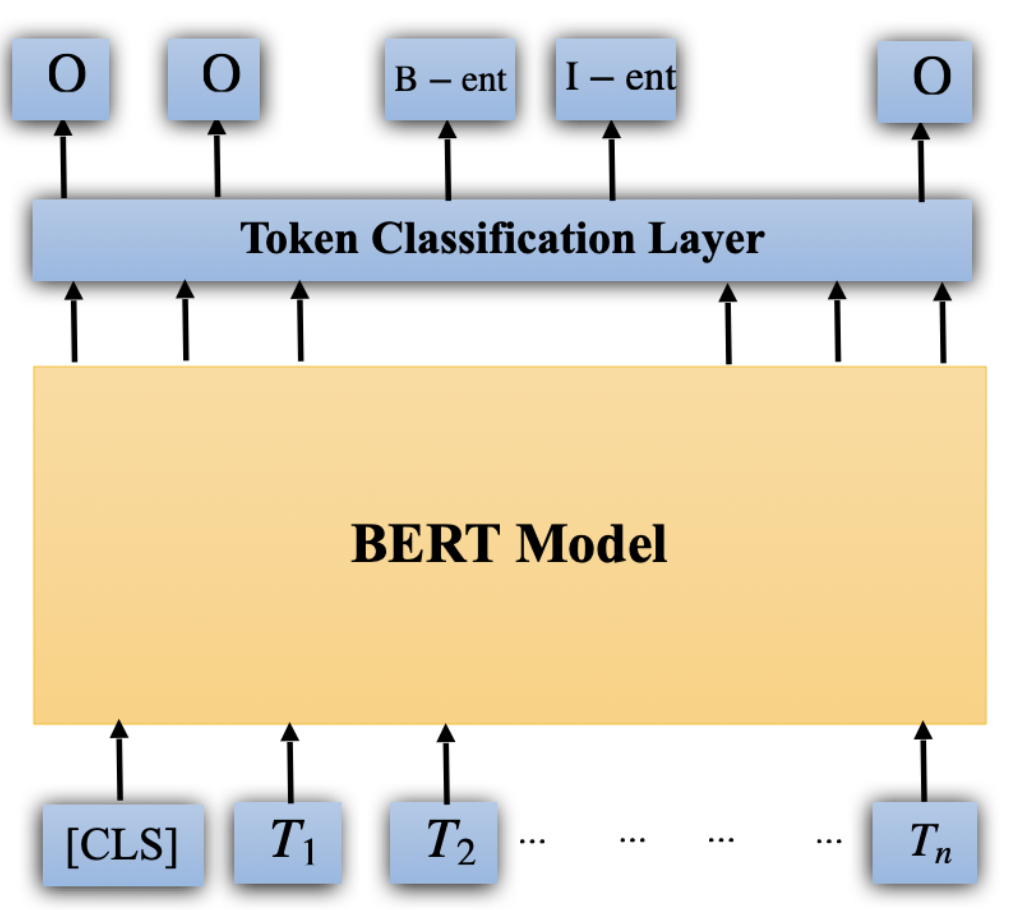}
\caption{The NER model used to extract the entity}
\label{fig:figure7}
\end{subfigure}
\hfill
\begin{subfigure}[b]{0.49\textwidth}
\centering
\includegraphics[trim = 2.5cm 21cm 12cm 3cm, clip=true,width=7cm,height=4cm]{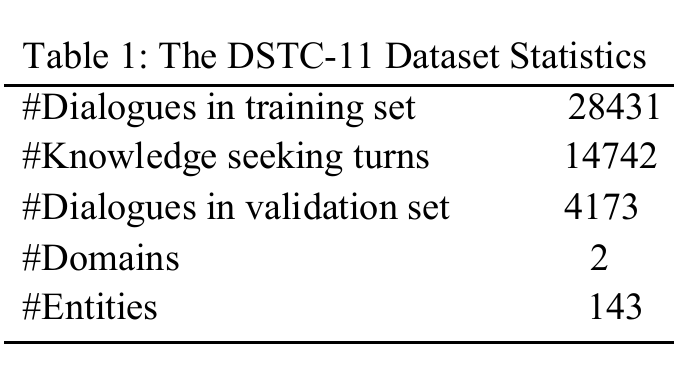}
\caption{Table 1}
\label{fig:table 1}
\end{subfigure}
\end{figure}

NER is a commonly used method to locate and classify named entities mentioned in unstructured text into predefined categories such as person names, organizations, locations etc. In the context of our tasks, the entity names are included in the conversation history almost all the time. Therefore, we define this problem as an NER which identifies the entities in the given conversation. We use the POS tagging to prepare the training dataset where each token is labeled as one of the followings, B-ent (Beginning entity), I-ent (Inside entity) and O (outside entity). Figure \ref{fig:figure6}  shows one example of dataset preparation. We get the true entity name from the entity name of the ground truth selected knowledge in training data. We fine-tune the BERT model using our training data to perform token classification. Fig \ref{fig:figure7} shows the model architecture. 

\subsection{Keyword Extraction for improved knowledge matching}\label{keyword extraction}

As shown in Fig \ref{fig:figure1}, we observe that we don’t need the whole conversation for the knowledge selection. We notice that there are a few keywords which are sufficient to find the relevant knowledge. These keywords are contained in the last few utterances of the conversation. Based on this observation, we propose to use a keyword extraction before knowledge matching. A model could not be trained to do this because of the limited time. However, we show an ablation study in Table \ref{table3} of Section \ref{results} to validate our claim.

\subsection{Dataset}\label{dataset}

We use the dataset provided in DSTC-11 challenge which is an augmented version of MultiWoz 2.1 \citep{zang-etal-2020-multiwoz} dataset . It includes newly introduced subjective knowledge-seeking turns. Table \ref{fig:table 1} shows the summary of the dataset. We train the baseline models for knowledge seeking turn detection and response generation using the provided dataset. However, for NER model training we prepare the dataset as explained in Section 3.2 using the ground truth from knowledge selection ground truth label.


\section{Implementation Details}\label{implementation details}

First, the baseline model for all three modules, namely KSTD, KS and RG, are trained. KSTC and KS use the DeBERTaV3 model \citep{he2021debertav3} while RG uses the BART model \citep{bart-base}. KSTC and RG are trained for 10 epochs and KS is trained for 3 epochs. It takes a total of $\sim$4 hours to train all three models on a single nvidia V100 GPU. For the NER training, the same DeBERTaV3 model \citep{he2021debertav3} is used to train it for 10 epochs on a single nvidia V100 GPU. All other hyper-parameters are kept as default values. The maximum token is trimmed to 256 for both history utterance and knowledge.

\section{Results}\label{results}

In this section, we present the experimental results obtained using the proposed entity retrieval methods using fuzzy logic and NER. In table \ref{table1} we show the accuracy of entity retrieval using proposed fuzzy logic, pre-trained NER model \citep{pretrained_ner} and fine-tuned NER. The proposed fuzzy logic always finds the entity name correctly. However it takes a long time since it uses n-gram comparison with all the entity names. Therefore, we use the NER to extract the entity name. The fine-tuned NER model achieves an accuracy of 72\% while being 7X more faster than fuzzy logic. 

\begin{table}
  \caption{Accuracy and inference speed comparison of the proposed fuzzy logic, a pretrained NER model[add ref] and fine-tuned NER model}
  \label{table1}
  \centering
  \begin{tabular}{lll}
    \toprule
    Method     & Accuracy(in \%) & Evaluation Time (in secs) \\
    \midrule
    Fuzzy logic & 100 & 1459\\
    Pre-trained-NER     & 65 & 220\\
    Fine-tuned NER     & 72 & 220 \textbf{($\sim$7X speedup)}\\
    \bottomrule
  \end{tabular}
\end{table}

In table \ref{table2} we show the ablation study of how many turns are enough to extract the entity name from the conversation history. We observe that using only the last 5 utterances are enough to find the entity name most of the time. A similar trend is observed for the fine-tuned NER. 

\begin{table}
\caption{Ablation study of how many turns are needed to find the entity name}
\label{table2}
\centering
\begin{tabular}{lll}
\toprule
\# dialogue turns     & Fuzzy logic(in \%) & Fine-tuned NER (in \%) \\
\midrule
1  & 15 & 11\\
2     & 84 & 56\\
5     & 98 & 66 \\
all & 100 & 72 \\
    
    \bottomrule
  \end{tabular}
\end{table}

Further, We perform an ablation study to find whether the proposed keyword extraction can improve the knowledge selection. To this end, we perform the knowledge selection by varying the number of dialogue turns. Table \ref{table3} shows that using only the last utterance leads to an improvement of $\sim$4\% in exact match accuracy as compared to when we use all the utterances. We also observe an improvement in the response generation bleu and rouge1 score with an increase in knowledge selection accuracy, as these three tasks are performed in sequential manner.

\begin{table}
  \caption{Preliminary results showing the effectiveness of keyword extraction.}
  \label{table3}
  \centering
  \begin{tabular}{lllll}
  
    \toprule & \multicolumn{2}{c}{Knowledge Selection} &  \multicolumn{2}{c}{Response Generation} \\
    \cmidrule(lr){2-3}
    \cmidrule(lr){4-5}
    
    \# dialogue turns & f1 & em\_accuracy(in \%) & bleu & rouge1 \\
    \midrule
    1 & 0.833 & 39.5 & 0.103 & 0.361 \\
    2 & 0.825 & 35.8 & 0.102 & 0.358 \\
    5 & 0.828 & 37.6 & 0.103 & 0.361\\
    all & 0.818 & 35.8 & 0.101 & 0.358\\
    
    \bottomrule
  \end{tabular}
\end{table}

\section{Future work}\label{future work}

Although this report shows that the NER based entity retrieval is much faster, it is still to reach the desired accuracy. Therefore, a future scope of work is to improve the accuracy of NER for this specific task. Another scope for improvement is with respect to the generalizability of the model. The training set provided in DSTC-11 has data only from the ‘hotel’ and ‘restaurant’ domain. Although the proposed method does not utilize this information, the model can be biased and hence may not generalize well to data from other domains like ‘flight’, ‘taxi’ etc. Hence, the present work can be extended to improve the generalizability of the model.

\section{Conclusion}\label{conclusion}
Knowledge grounded dialogue systems are the need of the hour, to make dialogue systems accurate and cohesive. This report finds that knowledge selection is the bottleneck for improving the performance of such a system. In this report the knowledge selection step is further broken down into an entity retrieval step followed by a knowledge matching step. To make the search faster, two variants of the entity retrieval method are proposed. The proposed fuzzy logic based entity retrieval achieves 100\% accuracy, while the NER based entity extraction makes the inference faster. Further, an ablation study shows that knowledge selection can be improved by performing keyword extraction.

\bibliographystyle{plainnat}
\bibliography{mybibfile}

\end{document}